\documentclass{article}

\usepackage{arxiv}

\usepackage[utf8]{inputenc} 
\usepackage[T1]{fontenc}    
\usepackage{hyperref}       
\usepackage{url}            
\usepackage{booktabs}       
\usepackage{amsfonts}       
\usepackage{nicefrac}       
\usepackage{microtype}      
\usepackage{lipsum}
\usepackage{graphicx}
\graphicspath{ {./images/} }

\usepackage{etoolbox}
\makeatletter
\patchcmd{\maketitle}{\@fnsymbol}{\@arabic}{}{}
\makeatother

\title{Suspicious Massive Registration Detection via Dynamic Heterogeneous Graph Neural Networks\thanks{This work is accepted in the AAAI Workshop on Deep Learning on Graphs 2021. We sincerely thank the reviewers for their constructive feedback, which we have incorporated into the current version of manuscript.} }

\author{
 Susie Xi Rao \\
  KOF Swiss Economic Institute \& \\ Systems Group\\
  ETH Zurich\\
  \texttt{raox@inf.ethz.ch}\\

   \And
  Shuai Zhang\\
   Systems Group\\
  ETH Zurich\\
  \texttt{shuazhang@inf.ethz.ch} \\
  
     \And
  Zhichao Han\\
   eBay China\\
  \texttt{zhihan@ebay.com} \\
  
       \And
  Zitao Zhang\\
   eBay China\\
  \texttt{zitzhang@ebay.com} \\
  
         \And
  Wei Min\\
   eBay China\\
  \texttt{wmin@ebay.com} \\

         \And
  Mo Cheng\\
   eBay China\\
  \texttt{mocheng@ebay.com} \\
  
         \And
  Yinan Shan\\
   eBay China\\
  \texttt{yshan@ebay.com} \\

         \And
  Yang Zhao\\
   eBay China\\
  \texttt{yzhao5@ebay.com} \\
  
  \And
  Ce Zhang\\
   Systems Group\\
  ETH Zurich\\
  \texttt{ce.zhang@inf.ethz.ch} \\
  
}

\begin{document}
\maketitle
\begin{abstract}
Massive account registration has raised concerns on risk management in e-commerce companies, especially when registration increases rapidly within a short time frame. To monitor these registrations constantly and minimize the potential loss they might incur, detecting massive registration and predicting their riskiness are necessary.
In this paper, we propose a \textbf{D}ynamic \textbf{H}eterogeneous \textbf{G}raph Neural Network  framework to capture suspicious massive \textbf{reg}istrations (\textbf{DHGReg}). We first construct a dynamic heterogeneous graph from the registration data, which is composed of a structural subgraph and a temporal subgraph. Then, we design an efficient architecture to predict suspicious/benign accounts. Our proposed model outperforms the baseline models and is computationally efficient in processing a dynamic heterogeneous graph constructed from a real-world dataset. In practice, the DHGReg framework would benefit the detection of suspicious registration behaviors at an early stage.
\end{abstract}


\section{Introduction}
Fraudulent buyer registrations may result in a large number of high-risk buyer accounts, which leave an opportunity for fraudulent purchase activity and may incur significant losses for customers. Today's fraudsters do not use single accounts to conduct fraud, but instead they tend to construct an army of risky accounts (e.g., bot armies). This can result in massive attacks on normal users with a range of tactics. These tactics include paying with stolen credit cards, chargeback fraud involving complicit cardholders, selling fake e-gift cards, or creating schemes to create large-scale layered fraud against multiple merchants\footnote{https://www.verifi.com/in-the-news/need-know-fraud-rings/ (last accessed: Dec 7, 2020).}. In this work, we aim to detect the massive registration of these risky accounts in a real-world e-commerce platform.

The existing solution for high-risk registration detection mainly relies on a rule-based model and manual review of \textit{individual} accounts. This approach has four downsides. Firstly, most such rules contain hard-coded known risk features, which are only a small subset of all features available for buyer accounts. Therefore, such rules may not have a high capture rate and are not adaptive to new fraud patterns. Secondly, building a rule-based model may come at the cost of suspending registrations, which can affect the normal business of customers. Thirdly, the reliance on manual review makes the process less scalable, and the capacity of a rule-based model is limited in dealing with a fraud ring attack. Lastly, and most importantly, the rule-based model only provides a viewpoint of individual accounts. Nonetheless, well-organized fraudsters usually sign up a massive amount of accounts to conduct their fraudulent behaviors on the platform. 

We observe two main characteristics of massive registration. Firstly, abusive accounts are usually interlinked. For example, they may share the same phone number or be registered with the same IP address. They naturally form a graph with heterogeneous nodes (e.g., accounts, IP address, email, phone number). Secondly, we want to capture suspicious massive registrations within a certain time frame. By studying the patterns of suspicious registrations, our business unit has discovered that because fraudsters tend to only abuse accounts when they are recently registered, the temporal dynamic is a critical factor for detection. In our effort to improve the efficacy and capability of highly risky massive registration detection, we have come up with a process solution based on a dynamic heterogeneous graph neural network (DHGReg). 

Learning from graphs with neural networks is gaining increasing popularity. In particular, graph neural network (GNN)~\cite{kipf2016semigcn, vaswani2017attention, hamilton2017inductive-sage} is powerful in capturing the graph structure and complex relations amongst nodes via message passing and agglomeration.  A GNN based system that detects massive registration is expected to deal with the challenges in (1) learning embedding representations of entities in a dynamic graph; (2) scaling to large graphs; and (3) utilizing neighborhood information in a local community in a dynamic heterogeneous transaction graph.

The proposed DHGReg makes sure that various types of entities are present in the registration graph and its temporality property. It operates on heterogeneous graphs, and the temporal information of graphs are also integrated during learning. It not only enjoys the benefits of general GNNs but can also effectively handle the heterogeneity and dynamics of real-world registration graphs.

Our key contributions in this work are:
\begin{itemize}
    \item We have designed a system DHGReg that models the dynamics of massive registration in a heterogeneous transaction graph.
    \item Our system outperforms the baseline models in detecting massive registration by 1.5\% to 3.6\%.
    \item Our system has performed efficiently in a transaction graph with up to 130,766 nodes and 288,700 edges (153,086 entity edges + 135,614 temporal edges).
\end{itemize}

\section{Related Work}\label{sec:lit}

In this section, we discuss previous works related to our method. Learning in heterogeneous graphs has gained interest in recent years, and several recent works aim to generalize the traditional graph convolutional network (GCN) and graph attention network (GAT) to heterogeneous graphs~\cite{hu2020hgt, wang2019han, hong2020attention, zhang2019hetgnn}. These models specify node types when constructing graphs and perform sampling over different types of nodes during message passing, which brings improvement when working with heterogeneous graphs.  The temporal dynamics are oftentimes investigated within a homogeneous graph setting. One representative work is DySAT~\cite{sankar2020dysat} that discusses an approach to learn deep neural representations on dynamic homogeneous graphs via self-attention networks. DySAT is applicable in cases where all entities in a graph have a dynamic perspective. However, in massive registration detection, we need to differentiate between two types of entities in the time dimension: (1) accounts being dynamically added/removed across time, (2) hard linking entities such as registration addresses and telephone numbers that stay static across time. 

It is shown in previous works on fraud detection~\cite{liu2018heterogeneous, wen2020asa, li2019spam} that heterogeneous graphs and GNN architectures designed specifically for them are powerful in capturing fraudulent patterns. Mining heterogeneous graphs in a dynamic setting has attracted growing interest in e-commerce-related applications. Shekhar et al.~\cite{shekhar2020entity} propose TIMESAGE for entity resolution, which learns entity representation from temporal weighted edges. Using temporal random walks, TIMESAGE also learns interaction sequences that evolve over time. TIMESAGE is tested for entity resolution, where transactions do not have an associated account ID, but instead each transaction is identifiable by IP address, email, and so forth. 

In our application, however, the use case is different. We do not aim at linking transactions and account registrations, but we are confronted with the challenge of detecting these suspicious accounts as early as possible. In addition, we aim at incorporating two types of nodes (static and dynamic nodes) as described above. Hence, we propose DHGReg to identify suspicious newly registered accounts even before any transaction has occurred. The linking entities in our heterogeneous graph are static nodes that can exist in every snapshot of the network. In contrast, newly registered accounts (dynamic nodes) can only appear after certain time periods.

\section{Research Question and Methodology}
In this section, we first present our research question, the problem definition of massive registration from the perspective of modeling a dynamic heterogeneous graph. Then, we explain the graph construction and label prediction in DHGReg.

\subsection{Research Question}
To efficiently detect massive account registration using graph level information, we aim to answer this question: how can we leverage the entities used in account registrations to preemptively predict suspicious registration conducted by potential fraudsters?

\begin{figure*}[ht]
    \centering
    \includegraphics[width=1\textwidth]{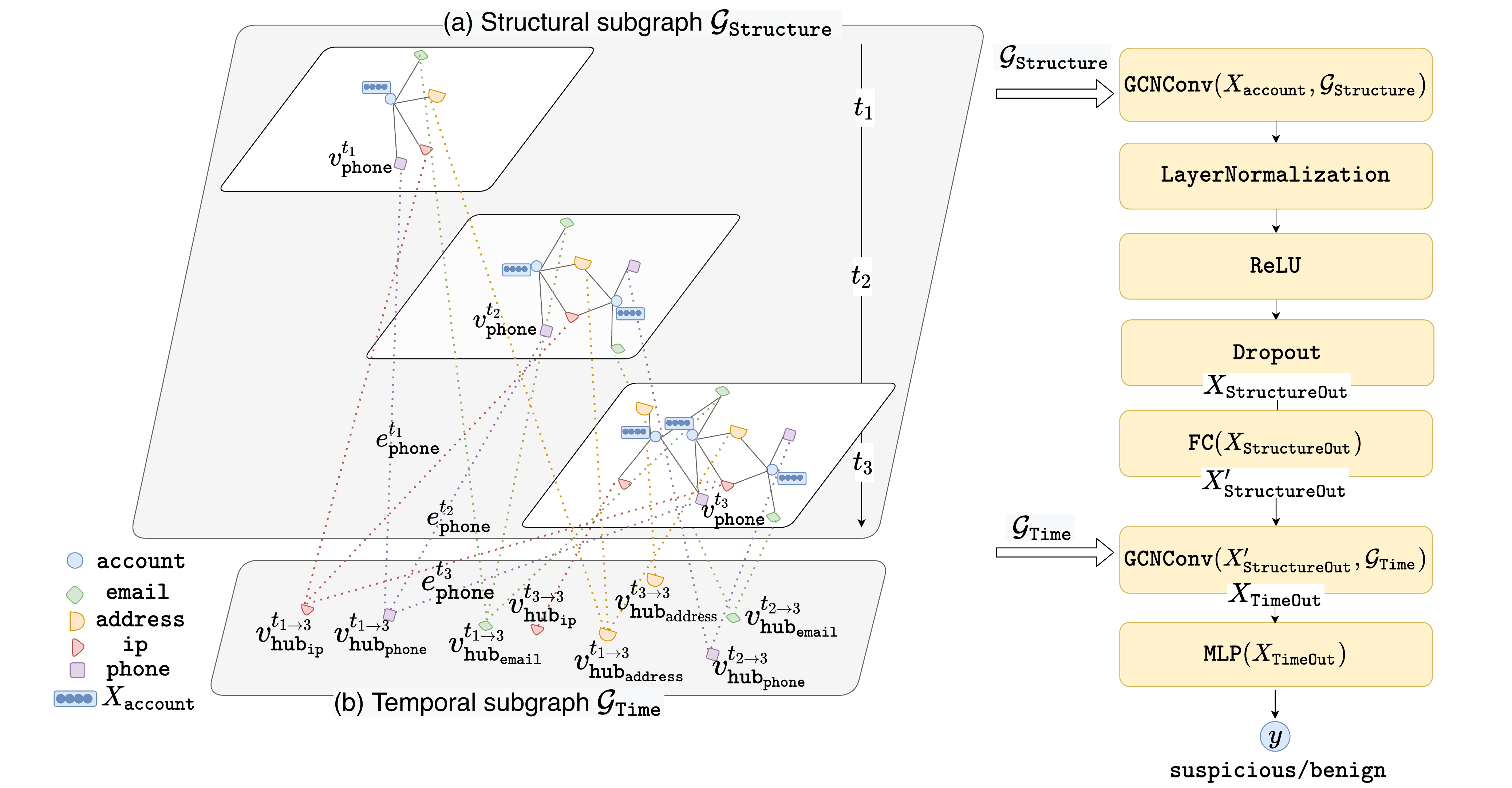}
    \caption{Dynamic Heterogeneous Graph Neural Network for Suspicious Massive Registration (DHGReg).}
    \label{fig:DHGReg}
\end{figure*}

\subsection{Problem Definition}\label{sec:problem}
Massive account registration refers to a process where a user or an organization can add a group of accounts simultaneously\footnote{See this illustration for an experience of massive user registration, https://hellohelp.gointegro.com/en/articles/1488462-massive-users-registration (last accessed: Nov. 9, 2020).}. This functionality is enabled on many e-commerce websites to speed up batch registration for a group of accounts. By providing the entities required in a registration template, a user can create within a short period several accounts. Apart from template-based operations, a bot can also be built to submit forms multiple times with generic information. In practice, a fraudster can make use of this functionality on an e-commerce website to register a set of accounts for multiple purposes: using stolen financial means, providing fake reviews, redeeming coupons and purchasing using an abusive gift card, etc. Oftentimes, these registrations share some common patterns of disguise.  

Think of the critical entities involved in massive account registration. A credit card might be stolen and used by one of the newly registered accounts. A suspicious account can be registered with a common shipping address such as a warehouse, which is a disguise of suspicious activity. These accounts are oftentimes registered using email addresses with spam patterns, also using telephone numbers that are listed as spam calls by third-party collaborators for risk detection. We can also detect a chain of illegal activities within an e-commerce platform: an account has been hacked and a fraudster registers a batch of new accounts to use the gift cards and financial instruments that are bundled with the hacked account. 

Furthermore, the time dimension is crucial in massive registration. Based on the manual analysis of the business unit on suspicious massive registrations, usually within three months after registration, fraudster activities occur. An effective system should be able to link suspicious entities used in the past or uncover suspicious patterns discovered in the past graph snapshots. Later, when such a system is deployed in production, it is expected to provide both feature level detection and graph level detection. Feature level detection is achievable via a rule-based system, while graph level detection is better modeled by an automatic model that learns the time-variant representations quickly and effectively.

Formally, a heterogeneous network is defined as a graph $\mathcal{G} = (\mathcal{V}, \mathcal{E}, \mathcal{A}, \mathcal{R})$, where node type mapping function is $\tau : \mathcal{V} \rightarrow \mathcal{A}$ and link type mapping function is $\phi : \mathcal{E} \rightarrow \mathcal{R}$. Each node $v \in \mathcal{V}$ has only one type $\tau(v) \in \mathcal{A}$, and each edge $e \in \mathcal{E}$ has only one type $\phi(e) \in \mathcal{R}$. Each node or edge can be associated with attributes, denoted by $X_{\tau({v})}$. In addition, each node or edge can be labeled, denoted by $y \in \mathcal{C}$. For each time $t$, an entity in $\mathcal{V}$, or a link in $\mathcal{E}$, can be removed from or added into the graph snapshot $\mathcal{G}_t$.

\subsection{Heterogeneous Graph Construction}
If we formulate massive registration detection as a binary classification problem in a transductive setting in a heterogeneous graph, we have the specification of the problem formulation as follows. 
In a heterogeneous transaction graph $\mathcal{G}$, $v \in \mathcal{V}$ has a type $\tau(v) \in \mathcal{A}$, where $\mathcal{A}:= \{\textit{account, address, ip, phone, email}\}$, referring to account, registration address, IP address, phone, email, respectively\footnote{For this study, we choose these attributes because they reflect typical patterns of massive registration activities as reported by our business unit. Other types of attributes such as device type are incorporated in the in-house risk detection system.}. If an account uses a linking entity in $\{$\textit{address, ip, phone, email}$\}$, we put an edge between the account node and that linking entity in the heterogeneous graph. Each $account$ node carries node attributes provided by a risk identification system. Each account ID is flagged as benign or suspicious. We use these flags as labels in our binary classification.

There are two ways (\textit{static} and \textit{dynamic}) to incorporate the time dimension into a heterogeneous graph using accounts and their linking entities. 

\subsubsection{Static}
One can either construct different static snapshots of a graph with all involving entities for each time $t$. This means a detection system has to process all snapshots $\{G_1,\dots,G_T\}$, where $T$ is the number of time steps. In each snapshot, a graph is constructed. This is the graph construction DySAT~\cite{sankar2020dysat} employs. 

However, as analyzed in Section~\ref{sec:lit}, DySAT is not suitable for our application because not all entities are appearing in every snapshot. If we break down the graph into snapshots, we will lose many time-dependent linkages and because many accounts have not existed and hence cannot be linked to any entity. Another reason is that newly registered accounts that are suspicious will only be abused within the $T$ weeks ($T$ is usually small, say 1-12). Consequently, we only need to encode the linking entities such as IP addresses and registration addresses for those periods. Hereafter, we propose an alternative of constructing the graph for our application and present it below.

\subsubsection{Dynamic}
Unlike treating a heterogeneous graph as a set of static snapshots, we unroll the time snapshots to incorporate nodes and edges that (dis)appear overtime in one graph. TIMESAGE~\cite{shekhar2020entity} has showcased is a similar design: a time-dependent network representation. We build a graph that can track entities that could be present or absent in all snapshots. Entities such as phone numbers and addresses are present in all snapshots, but accounts can only exist after certain snapshots, i.e., after the accounts have been created.

We have two main components in our dynamic heterogeneous transaction network, a structural subgraph (Figure~\ref{fig:DHGReg} (a)), and a temporal subgraph (Figure~\ref{fig:DHGReg} (b)). Here we discuss the design of these two network components and the intuitions behind the design. The structural subgraph is designed to reflect the linkages among various types of entities. The nodes are accounts and attributes used in account registration. For instance, if two accounts are registered using the same IP, an edge is added from account 1 to this IP, and another edge is added from account 2 to this IP as well. Consequently, account 1 and account 2 are linked via this IP in this way (see Figure~\ref{fig:DHGReg} (a)). Hence, the structural subgraph captures the relationships among the entities and allows us to uncover the patterns in account registration. Now let us talk about the temporal subgraph that builds on top of the structural components in each time $t$. For each time $t$, we observe a structural subgraph constructed as we describe above. Then, we add temporal edges from the structural nodes to a node called $v_{hub}$. These structural nodes in different timestamps represent the identical entities in each time $t$. We also index $v_{hub}$ with the timestamp(s) when it appears. In the example shown in Figure~\ref{fig:DHGReg}, the nodes $v_{hub_{phone}}^{t_{1}}$, $v_{hub_{phone}}^{t_{2}}$, $v_{hub_{phone}}^{t_{3}}$ are connected to a node $v_{hub_{phone}}^{t_{1\rightarrow3}}$ via the temporal edges $e_{phone}^{t_{1}}$, $e_{phone}^{t_{2}}$, and $e_{phone}^{t_{3}}$. From the $v^t$ nodes to the $v_{hub}$ node we have a star graph that represents if an entity has appeared in time $t$ or not. We denote the unrolled dynamic heterogeneous graph as $\mathcal{G}_T$, where all the edges and nodes appearing from $\{1, \dots, T\}$ are present. $\mathcal{G}_T$ is composed of $\mathcal{G}_{Structure}$ and $\mathcal{G}_{Time}$.

\begin{table}[!t]
    \centering
    \begin{tabular}{l|l}
    \toprule
         \textbf{Notation} & \textbf{Description}  \\
    \midrule
         $T$ & the number of time steps \\
         \midrule
         $\mathcal{G}_T$ & a dynamic heterogeneous graph from timestamps $\{1, \dots, T\}$ \\
         \midrule
         $\mathcal{G}_{Structure}$ & structural subgraph in Figure~\ref{fig:DHGReg} (a) \\
         \midrule
         $\mathcal{G}_{Time}$ & temporal subgraph in Figure~\ref{fig:DHGReg} (b) \\
         \midrule
         $X_{account}$ & account level features \\
         \midrule
         $X_{StructureOut}$ & the output of structural message passing \\
         \midrule
         $X^{\prime{}}_{StructureOut}$ & the output of FC transformation of $X_{StructureOut}$ \\
         \midrule
         $X_{TimeOut}$ & the output of structural message passing \\
         \midrule
         $e_i^t$ & an edge in each time $t$ for type $i$, e.g., $e_{phone}^{t_1}$ \\
         \midrule
         $v_i^t$ & a node in each time $t$ for type $i$, e.g., $v_{phone}^{t_1}$ \\
         \midrule
         $v_{{hub}_i}^{{t_{i\rightarrow j}}}$ & a hub node in from $t_i$ to $t_j$ for type $i$, e.g., $v_{hub_{phone}}^{t_{1 \rightarrow 3}}$ \\
    \bottomrule 
    
    \end{tabular}
    \vspace{3mm}
    \caption{Notations.}
    \label{tab:notations}
\end{table}

\begin{table}[!t]
    \centering
    \begin{tabular}{c|r|c|r}
        \toprule
        \textbf{Node Type} & \textbf{Count} & \textbf{Edge Type} & \textbf{Count} \\
        \midrule
         Account ID & 111,691 & Temporal Edge & 135,614\\
         Email & 7,221 & Account ID - Email & 29,217 \\
         Address & 6,762 & Account ID - Address & 104,719\\
         Phone & 4,958 & Account ID - Phone & 18,542 \\
         IP Address & 134 & Account ID - IP Address & 608 \\
         \midrule
         TOTAL & 130,766 & TOTAL & 288,700 \\
        \bottomrule
        
    \end{tabular}
        \vspace{3mm}

    \caption{Count of Node and Edge Types in our Heterogeneous Graph.}
    \label{tab:node-edge-count}
\end{table}

\subsection{DHGReg Architecture}
\label{sec:architecture}
Now we describe in detail how we implement GCN convolutional layers on the structural and temporal subgraphs. We denote the structural and temporal subgraphs as $\mathcal{G}_{Structure}$ and $\mathcal{G}_{Time}$, respectively. For each subgraph, we implement a GCN convolution layer to learn the message passing between the nodes.  
\subsubsection{The Structural Message Passing} We denote the account level feature as $X_{account}$. The input to the GCN layer on the structural subgraph is $\mathcal{G}_{Structure}$ and $X_{account}$. The other types of nodes do not have initial feature values and are initialized with zero vectors. Then, the output is layer-normalized and fed into a nonlinear transformation using \textit{ReLU} as the activation function. Finally, a \textit{Dropout} layer is applied to regularize and avoid overfitting. Let us denote the output of the structural message passing as $X_{StructureOut}$. Then, we use a feedforward layer (FC in Figure~\ref{fig:DHGReg}) to connect the structural message passing and the temporal message passing. The input of FC is $X_{StructureOut}$, the output $X^{\prime{}}_{StructureOut}$.

\subsubsection{The Temporal Message Passing} 
The input of the temporal GCN convolution is (1) $X^{\prime{}}_{StructureOut}$, the output after the FC transformation of $X_{StructureOut}$,  and (2) $\mathcal{G}_{Time}$, the temporal subgraph. This means we only take the nodes of type $i$, which are linked by the temporal edges $e_i^t$ in time $t$, where $t \in \{1, \dots, T\}$, $i \in \{$\textit{address, ip, phone, email}$\}$. We denote the output from the temporal message passing as $X_{timeOut}$.

\subsubsection{Prediction} 
Then, the output $X_{timeOut}$ is fed into a feedforward connected network. We then apply dropout, layer normalization, and \textit{ReLU} transformation before a label is calculated (this part is summarized under MLP in Figure~\ref{fig:DHGReg}). The loss function is a cross entropy of the true label, and the probability score is calculated by \textit{softmax}.

\begin{table*}[!t]
\centering

\resizebox{\linewidth}{!}{\begin{tabular}{cc||c||cccccccccc}
\toprule
\textbf{Model}    &  \textbf{Average Precision (AP)}            & \textbf{Time Cost (s/epoch)} & \textbf{n\_layer} & \textbf{n\_hid} & \textbf{dropout} & \textbf{optimizer} & \textbf{learning rate} & \textbf{beta} & \textbf{max\_epochs} & \textbf{patience} & \textbf{n\_heads }\\
\midrule
MLP      & 0.7890$\pm$0.0042     &   0.06   & \{4, 8, 12\}& 256 &\{0.1, 0.25, 0.5\}   &adamw &0.001& (0.9, 0.999) & 2048 & 64 &-\\
GCN      & 0.8093$\pm$0.0047      &  0.30  & \{4, 8, 12\}& 256 &\{0.1, 0.25, 0.5\}   &adamw &0.001& (0.9, 0.999) & 2048 & 64&-\\
GAT      & 0.8068$\pm$0.0034       &  0.40 & \{4, 8, 12\}& 256 &\{0.1, 0.25, 0.5\}   &adamw &0.001& (0.9, 0.999) & 2048 & 64&8\\
DHGReg (ours) & \textbf{0.8251$\pm$0.0056} & 0.40 & \{4, 8, 12\}& 256 &\{0.1, 0.25, 0.5\}   &adamw &0.001& (0.9, 0.999) & 2048 & 64&-\\
\bottomrule
\end{tabular}}
    \vspace{3mm}

\caption{Experiment Results and Hyperparameters of DHGReg and its Baselines.}
\label{tab:result}

\end{table*}

\section{Experiments, Evaluation and Discussions}
In this section, we introduce our dataset and its preprocessing, show the experimental setup for DHGReg, evaluate the performance of our model, and discuss our findings.

\subsection{Dataset, Preprocessing and Evaluation Metric}
\label{sec:data}

To model massive registration detection, we construct a heterogeneous graph by transforming the registration records from an e-commerce platform. The records were selected from the account registrations from September to December in 2019. Account features and linking entities were compiled 24 hours after an account had been created. Account features include risk features generated by an in-house risk detection system. The dimension of risk features is 264. Account features are composed of the following attributes: (1) registration profiles attributes: such as registration email, user name, text patterns in an address, IP location; (2) registration behavior patterns: registration click stream summary, traffic information; (3) transaction patterns: early purchasing patterns, such as velocity, price, and item categories.  

Our dataset was sampled from the real-time account creation logs. We utilize METIS~\cite{karypis1998fast} which partitions the graph with connectivity guarantees and controls the sparsity of the local graphs.
The expectation of edge connectivity is set to 512. The parameter $k$ denoting a $k$-way partition is calculated by the number of nodes in a heterogeneous graph divided by 512. For the time dimensions, we slice the time window to be a week and create the edges based on these time sliding windows. 
In Table~\ref{tab:node-edge-count}, we summarize the node and edge counts for our dataset.

Out of 111,691 account registrations, we have 63,161 accounts with suspicious flags. These labels are either automatically generated based on rule-based filtering on transaction behaviors (riskiness, payment rejection, chargeback, abusive buyers, etc) or based on manual annotation deducing from the registration rules. One of these registration rules could be, for instance, an IP address that has over 500 registrations should be flagged as suspicious before further operation. 

As our dataset is balanced with 56.55\% suspicious registrations, we use average precision (AP), because it is a popular evaluation metric in risk management. It computes the average precision value for the recall value over 0 to 1, i.e., the area under the precision-recall curve. AP is interpreted as the mean of the precision scores after each relevant sample is retrieved~\cite{zhang2009averageprecision}. We split the training and testing sets in the ratio of 80\% and 20\% based on week numbers. 


\subsection{Implementation Details: DHGReg and its baselines}
\label{sec:implementation}
We have discussed in Section~\ref{sec:lit} why DySAT~\cite{sankar2020dysat} and TIMESAGE~\cite{shekhar2020entity} do not apply in detecting massive registration. Therefore, we select our baseline models to be multilayer perceptron (MLP) and simple GNN models such as GCN and GAT which were originally designed to model homogeneous graphs. All models are implemented in PyTorch and trained using batch learning. Here below, we report the detailed settings of each model and its hyperparameters in Table~\ref{tab:result} based on the model performance on the validation set.

\begin{itemize}
    \item MLP. A feedforward neural network with only account level features $X_{account}$ as input. It uses \textit{ReLU} as the activation function.

    \item GCN. To train a GCN/GAT model, we still construct a heterogeneous graph as described in Section~\ref{sec:data}. The goal of running GCN and GAT is to see if methods like GCN and GAT, which were originally developed for homogeneous graphs, can be outperformed by architectures designed specifically for heterogeneous graphs such as DHGReg. 
    
    \item GAT. Comparing with GCN, which aggregates the mean features from the one-hop neighbors, GAT uses an attention mechanism over the features of one-hop neighbors. 

\end{itemize}


\subsection{Results and Analysis}

We report the AP scores on DHGReg and its baselines in Table~\ref{tab:result}. Here we report the best performing DHGReg model after hyperparameter tuning. Note that we primarily tune two hyperparameters \texttt{n\_layer} and \texttt{dropout}. The set of hyperparameters in the best model are $\texttt{n\_layer} = 4$, $\texttt{dropout}=0.1$, the remaining hyperparameters see Section~\ref{sec:implementation}. For every baseline, we only report its performance on the corresponding setup with identical hyperparameters. Comparing the results across models, we make observations as follows:
\begin{enumerate}
    \item DHGReg outperforms MLP by 3.61\%, GCN by 1.58\%, and GAT by 1.83\%.
    \item GNN based methods outperform non-GNN based baseline MLP. This signifies the benefit of incorporating graph-based structure into massive registration detection.
    \item Our proposed architecture DHGReg incorporates the temporal and structural aspects from various snapshots and shows its capability to generate better predictions. 
\end{enumerate}

To benchmark the training, we also report the time cost of DHGReg and its baselines in Table~\ref{tab:result}. We see that DHGReg, which uses four GCN convolution layers on a heterogeneous dynamic graph, runs as fast as a four-layer GAT network on the same input graph. Comparing the time cost of GCN and DHGReg, we see that the GCN convolution designed for the temporal subgraph adds an affordable cost in computation, in exchange for a better performance in predicting massive registration in our implementation.
We can use the massive registration predictions in speeding up the daily routine of the business unit. Detected highly risky accounts have been sent to the internal team for manual review, and more than 90\% of the accounts have been confirmed as massive registration accounts, and the majority of these accounts have abusive buying activities that lead to either financial loss or reputation damage. To protect the integrity of the marketplace, these accounts have been suspended.

\subsection{Discussions}
We are aware of the fact that the graph size for $\mathcal{G}_{Time}$ increases linearly as the number of time steps $T$ grows. There might be discussions on the scalability in real-life detection, when $T$ is extremely large. However, in our application scenario of massive registration detection, we aim at capturing the newly registered accounts and their activities as early as possible. Ideally, even before the first transactions of these accounts take place, we block the suspicious accounts for further investigation. Only in this way, we can preemptively control risks and avoid the interruption of normal activities on our platform. Consequently, when deploying such a detection system, we will make the $T$, e.g., the number of weeks, around 10-12 weeks. 

Certainly, there is a possibility that fraudters register many accounts for longer-term usage. They might invest in certain accounts that appear to be benign and conduct normal activities to avoid risk surveillance in the short run. In this case, following the first 10-12 weeks of account activities will exempt these accounts from being suspicious. Because the activities later on are relatively loosely connected with the registration patterns of the corresponding accounts, it needs another type of modelling to capture their fraudulent activities. Typically, to capture fraudulent activities conducted by these accounts, fraud detection should be conducted on the transaction level, see our work on xFraud~\cite{rao2020xfraud} for more details.

\section{Conclusion and Future Work}

We present an approach to effectively detecting massive account registration in a dynamic fashion. The DHGReg framework constructs a dynamic heterogeneous graph by adding structural and temporal edges into one single graph. Then, DHGReg predictor discovers suspicious accounts by combining structural GCN convolutions and temporal GCN convolutions. 

The paper is our first attempt to design a graph-based machine learning system in fraudulent massive registration detection. For future work, we have the following working directions: (1) We can implement DHGReg in an incremental setting where we append new registrations generated weekly and test how the model performs. (2) We can apply the framework to other risk scenarios with a dynamic aspect, such as fraud transaction detection in a dynamic graph. (3) Instead of GCN convolutions, we can use complex graph convolutional layers such as RGCN~\cite{schlichtkrull2018modeling}, GAT or APPNP\footnote{RGCN: Relational GCN, APPNP: Approximate Personalized Propagation of Neural Predictions.}~\cite{klicpera2018predict}. It would also be intriguing to design a DHGReg predictor using heterogeneous graph neural network models instead of two isolated network representations for structural and temporal representations. This could make the network representations more compact and more expressive on entities.

\bibliographystyle{unsrt}  
\bibliography{references} 

\begin{thebibliography}{10}

\bibitem{kipf2016semigcn}
Thomas~N Kipf and Max Welling.
\newblock Semi-supervised classification with graph convolutional networks.
\newblock {\em arXiv preprint arXiv:1609.02907}, 2016.

\bibitem{vaswani2017attention}
Ashish Vaswani, Noam Shazeer, Niki Parmar, Jakob Uszkoreit, Llion Jones,
  Aidan~N Gomez, {\L}ukasz Kaiser, and Illia Polosukhin.
\newblock Attention is all you need.
\newblock In {\em Advances in neural information processing systems}, pages
  5998--6008, 2017.

\bibitem{hamilton2017inductive-sage}
Will Hamilton, Zhitao Ying, and Jure Leskovec.
\newblock Inductive representation learning on large graphs.
\newblock In {\em Advances in neural information processing systems}, pages
  1024--1034, 2017.

\bibitem{hu2020hgt}
Ziniu Hu, Yuxiao Dong, Kuansan Wang, and Yizhou Sun.
\newblock Heterogeneous graph transformer.
\newblock In {\em Proceedings of The Web Conference 2020}, pages 2704--2710,
  2020.

\bibitem{wang2019han}
Xiao Wang, Houye Ji, Chuan Shi, Bai Wang, Yanfang Ye, Peng Cui, and Philip~S
  Yu.
\newblock Heterogeneous graph attention network.
\newblock In {\em The World Wide Web Conference}, pages 2022--2032, 2019.

\bibitem{hong2020attention}
Huiting Hong, Hantao Guo, Yucheng Lin, Xiaoqing Yang, Zang Li, and Jieping Ye.
\newblock An attention-based graph neural network for heterogeneous structural
  learning.
\newblock In {\em AAAI}, pages 4132--4139, 2020.

\bibitem{zhang2019hetgnn}
Chuxu Zhang, Dongjin Song, Chao Huang, Ananthram Swami, and Nitesh~V Chawla.
\newblock Heterogeneous graph neural network.
\newblock In {\em Proceedings of the 25th ACM SIGKDD International Conference
  on Knowledge Discovery \& Data Mining}, pages 793--803, 2019.

\bibitem{sankar2020dysat}
Aravind Sankar, Yanhong Wu, Liang Gou, Wei Zhang, and Hao Yang.
\newblock Dysat: Deep neural representation learning on dynamic graphs via
  self-attention networks.
\newblock In {\em Proceedings of the 13th International Conference on Web
  Search and Data Mining}, pages 519--527, 2020.

\bibitem{liu2018heterogeneous}
Ziqi Liu, Chaochao Chen, Xinxing Yang, Jun Zhou, Xiaolong Li, and Le~Song.
\newblock Heterogeneous graph neural networks for malicious account detection.
\newblock In {\em Proceedings of the 27th ACM International Conference on
  Information and Knowledge Management}, pages 2077--2085, 2018.

\bibitem{wen2020asa}
Rui Wen, Jianyu Wang, Chunming Wu, and Jian Xiong.
\newblock Asa: Adversary situation awareness via heterogeneous graph
  convolutional networks.
\newblock In {\em Companion Proceedings of the Web Conference 2020}, pages
  674--678, 2020.

\bibitem{li2019spam}
Ao~Li, Zhou Qin, Runshi Liu, Yiqun Yang, and Dong Li.
\newblock Spam review detection with graph convolutional networks.
\newblock In {\em Proceedings of the 28th ACM International Conference on
  Information and Knowledge Management}, pages 2703--2711, 2019.

\bibitem{shekhar2020entity}
Shubhranshu Shekhar, Deepak Pai, and Sriram Ravindran.
\newblock Entity resolution in dynamic heterogeneous networks.
\newblock In {\em Companion Proceedings of the Web Conference 2020}, pages
  662--668, 2020.

\bibitem{karypis1998fast}
George Karypis and Vipin Kumar.
\newblock A fast and high quality multilevel scheme for partitioning irregular
  graphs.
\newblock {\em SIAM Journal on scientific Computing}, 20(1):359--392, 1998.

\bibitem{zhang2009averageprecision}
Ethan Zhang and Yi~Zhang.
\newblock In Ling Liu and M~Tamer {\'O}zsu, editors, {\em Encyclopedia of
  database systems}, volume~6, chapter Average Precision. Springer New York,
  NY, USA, 2009.

\bibitem{rao2020xfraud}
Susie~Xi Rao, Shuai Zhang, Zhichao Han, Zitao Zhang, Wei Min, Zhiyao Chen,
  Yinan Shan, Yang Zhao, and Ce~Zhang.
\newblock xfraud: Explainable fraud transaction detection on heterogeneous
  graphs.
\newblock {\em arXiv preprint arXiv:2011.12193}, 2020.

\bibitem{schlichtkrull2018modeling}
Michael Schlichtkrull, Thomas~N Kipf, Peter Bloem, Rianne Van Den~Berg, Ivan
  Titov, and Max Welling.
\newblock Modeling relational data with graph convolutional networks.
\newblock In {\em European Semantic Web Conference}, pages 593--607. Springer,
  2018.

\bibitem{klicpera2018predict}
Johannes Klicpera, Aleksandar Bojchevski, and Stephan G{\"u}nnemann.
\newblock Predict then propagate: Graph neural networks meet personalized
  pagerank.
\newblock {\em arXiv preprint arXiv:1810.05997}, 2018.

\end{thebibliography}

\end{document}